\documentclass[10pt,twocolumn,letterpaper]{article}

\usepackage{iccv}
\usepackage{times}
\usepackage{epsfig}
\usepackage{graphicx}
\usepackage{amsmath}
\usepackage{amssymb}
\usepackage{algorithm} 
\usepackage{multirow} 
\usepackage{latexsym}
\usepackage{algpseudocode}
\usepackage{algorithmicx} 

\usepackage[breaklinks=true,bookmarks=false]{hyperref}

\iccvfinalcopy 

\def\iccvPaperID{11132} 

\def\TJ(#1){\textcolor{blue}{TJ: #1}}
\def\YH(#1){\textcolor{red}{YH: #1}}

\ificcvfinal\fi

\begin{document}

\title{Unsupervised Image Generation with Infinite Generative Adversarial Networks}


\author{
Hui Ying$^1$,\qquad
He Wang$^2$,\qquad
Tianjia Shao$^1$\thanks{Corresponding author. The authors from Zhejiang University are affiliated with the State Key Lab of CAD\&CG.},\qquad
Yin Yang$^3$,\qquad
Kun Zhou$^1$ \\
$^1$Zhejiang University\qquad
$^2$University of Leeds\qquad
$^3$Clemson University \\
{\tt\small huiying@zju.edu.cn,
H.E.Wang@leeds.ac.uk,
tjshao@zju.edu.cn,
yin5@clemson.edu,
kunzhou@acm.org}
}

\maketitle

\begin{abstract}
Image generation has been heavily investigated in computer vision, where one core research challenge is to generate images from arbitrarily complex distributions with little supervision. Generative Adversarial Networks (GANs) as an implicit approach have achieved great successes in this direction and therefore been employed widely. However, GANs are known to suffer from issues such as mode collapse, non-structured latent space, being unable to compute likelihoods, etc. In this paper, we propose a new unsupervised non-parametric method named mixture of infinite conditional GANs or MIC-GANs, to tackle several GAN issues together, aiming for image generation with parsimonious prior knowledge. Through comprehensive evaluations across different datasets, we show that MIC-GANs are effective in structuring the latent space and avoiding mode collapse, and outperform state-of-the-art methods. MIC-GANs are adaptive, versatile, and robust. They offer a promising solution to several well-known GAN issues. Code available: \url{github.com/yinghdb/MICGANs}.
\end{abstract}

\section{Introduction}
GANs have achieved great successes in a fast-growing number of applications~\cite{jabbar2020survey}. The success lies in their ability to capture complex data distributions in an unsupervised, non-parametric and implicit manner~\cite{goodfellow2017nips}. Yet, such ability comes with limitations, such as mode collapse. Despite a range of methods attempting to address these issues, they are still open. This motivates our research aiming to mitigate several limitations collectively including mode collapse, unstructured latent space, and being unable to compute likelihoods, which we hope will facilitate follow-up GAN research and broaden their downstream applications.


GANs normally consist of two functions: a generator and a discriminator. In image generation, the discriminator distinguishes between real and generated images, while the generator aims to fool the discriminator by generating images that are similar to real data. The widely known mode collapse issue refers to the generator's tendency to only generate similar data which aggregate around one or few modes in a multi-modal data distribution, e.g., only generating cat images in a cat/dog dataset. There has been active research in \textit{distribution matching} to solve/mitigate mode collapse~\cite{MMDGAN17,fGAN16,alphaGAN17,VEEGAN17}, which essentially explicitly/implicitly minimizes the distributional mismatch between the generated and real data. In parallel, it is found that latent space structuring can also help, e.g. by introducing conditions~\cite{cGAN14}, noises~\cite{styleGAN19}, latent variables~\cite{infoGAN16} or latent structures~\cite{deliGAN17}. In comparison, latent space structuring does enable more downstream applications such as controlled image generation, but they normally require strong prior knowledge of the data/latent space structure, such as class labels or the cluster number in the data or the mode number in the latent space. In other words, they are either supervised, or unsupervised but parametric and prescribed.

We simultaneously tackle the latent space structure and mode collapse by proposing a new, \textit{unsupervised} and \textit{non-parametric} method, mixture of infinite conditional GANs or MIC-GANs. Without loss of generality, we assume an image dataset contains multiple (unlabelled) clusters of images, with each cluster naturally forming one mode. Instead of making a GAN avoid mode collapse, we make use of it, i.e. exploiting GAN's mode collapse property, to let one GAN cover one mode so that we can use multiple GANs to capture all modes. Next, doing so naturally brings the question of how many GANs are needed. Instead of relying on the prior knowledge~\cite{benyosef2018gaussian,deliGAN17}, we aim to learn the number of GANs needed from the data. In other words, MIC-GANs model the distribution of an infinite number of GANs. Meanwhile, we also construct a latent space according to the data space by letting each GAN learn to map one latent mode to one data mode.
Since there can be an infinite number of modes in the data space, there are also the same number of modes in the latent space, each associated with one GAN. The latent space is then represented by a \textit{convex combination} of GANs and is therefore structured. 

To model a distribution of GANs, our first technical novelty is a new Bayesian treatment on GANs, with a family of non-parametric priors on GAN parameters. Specifically, we assume an infinite number of GANs in our reservoir, so that for each image, there is an optimal GAN to generate it. This is realized by imposing a Dirichlet Process~\cite{Ferguson_1973} over the GAN parameters, which partitions the probabilistic space of GAN parameters into a countably infinite set where each element corresponds to one GAN. The image generation process is then divided into two steps: first choose the most appropriate GAN for an image and then generate the image using the chosen GAN.

Our second technical novelty is a new hybrid inference scheme. Training MIC-GANs is challenging due to the infinity nature of DP. 
Not only do we need to estimate how many GANs are needed, we also need to compute their parameters.%
Some specific challenges include: 1) unable to compute likelihoods from GANs (a fundamental flaw of GANs)~\cite{Eghbalzadeh2017Likelihood}; 2) lack of an explicit form of GAN distributions; 3) prohibitive computation for estimating a potentially infinite number of GANs. These challenges are beyond the capacity of existing methods. We therefore propose a new hybrid inference scheme called Adversarial Chinese Restaurant Process. 

MIC-GANs are unsupervised and non-parametric. They automatically learn the latent modes and map each of them to one data mode through one GAN. MIC-GANs not only avoid mode collapse, but also enable controlled image generation, interpolation among latent modes, and a systematic exploration of the entire latent space. Through extensive evaluation and comparisons, we show the superior performance of MIC-GANs in data clustering and generation.


\section{Related Work}

\paragraph{Mode Collapse in GANs}
GANs often suffer from mode collapse, where the generator learns to generate samples from only a few modes of the true distribution while missing many other modes. To alleviate this problem, researchers have proposed a variety of methods such as incorporating the minibatch statistics into the discriminator~\cite{BatchDiscri16}, adding regularization~\cite{ModeRegu17,repelling17}, unrolling the optimization of the discriminator~\cite{MetzPPS17}, combining a Variational Autoencoder (VAE) with GANs using variational inference~\cite{alphaGAN17}, using multiple discriminators~\cite{MultiAdv17}, employing the Gaussian mixture as a likelihood function over discriminator embeddings~\cite{MDGAN19}, and applying improved divergence metrics in the loss of discriminator~\cite{wgan17,WGANGP17,LSGAN17,MiyatoKKY18,fGAN16}. Other methods focus on minimizing the distributional mismatch between the generated and real data. For example, VEEGAN~\cite{VEEGAN17} introduces an additional reconstructor network to enforce the bijection mapping between the true data distribution and Gaussian random noise. MMD GAN~\cite{MMDGAN17} is proposed to align the moments of two distributions with generative neural networks. Most of existing methods essentially map one distribution (often Gaussian or uniform) to a data distribution with an arbitrary number of modes. This is an extremely challenging mapping to learn, leading to many issues such as convergence and inablility to learn complex distributions \cite{OdenaOpen19}.  Rather than avoiding mode collapse, we \textit{exploit} it by letting one GAN learn one mode in the data distribution (assuming one GAN can learn one mode), so that we can use multiple GANs to capture all modes in the data distribution. This not only naturally avoids mode collapse but leads to more structured latent space representations.

\vspace{-10 pt}
\paragraph{Latent Space Structure in GANs}
Early GANs focus on mapping a whole distribution (e.g. uniform or Gaussian) to a data distribution. Since then, many efforts have been made to structure the latent space in GANs, so that the generation is controllable and the semantics can be learned. Common strategies involve introducing conditions~\cite{ModeSeek19,cGAN14,cGANProject18,ACGAN17,DiversityCGAN19}, latent variables~\cite{infoGAN16}, multiple generators~\cite{multiagent18,coupledGAN16}, noises~\cite{styleGAN19,styleGAN2} and clustering~\cite{mukherjee2019clustergan}. Recent approaches also employ mixture of models (e.g. Gaussian mixture models) to explicitly parameterize the latent space~\cite{benyosef2018gaussian,deliGAN17}. However, these methods usually require strong prior knowledge, e.g. class labels, the cluster number in the data distribution and the mode number in the latent space, with prescribed models to achieve the best performance. In this paper, we relax the requirement of any prior knowledge of the latent/data space. Specifically, MIC-GANs are designed to learn the latent modes and the data modes simultaneously and automatically. This is realized by actively constructing latent modes while establishing a one-to-one mapping between latent modes and data modes, where each GAN learns one mapping. Consequently, the latent space is structured by a convex combination of GANs. DMGAN~\cite{khayatkhoei2018disconnected} is the most similar work to ours, which employs multiple generators to learn distributions with disconnected support without prior knowledge. In contrast, MIC-GANs neither impose any assumption on the connectivity of the support,  nor require multiple generators. Besides, MIC-GANs have a strong clustering capability for learning the latent modes.
An alternative approach is to use Variational Autoencoder (VAE) which can structure the latent space (i.e., a single Gaussian or mixed Gaussians) during learning~\cite{VarPosterDirichlet20,VARDeepEmbed17, DirichletvAE20,AutoVarBayes14,SBVAE17,VAmpPriorVAE18}, but they often fail to generate images with sharp details. We therefore focus on GANs.

\section{Methodology}
\subsection{Preliminary}
Given image data $X$, a GAN can be seen as two distributions $G(X|\theta^g, Z)$ and $D([0,1]|\theta^d, X)$, with $\theta=[\theta^g, \theta^d]$ being the network weights and $Z$ being drawn from a distribution, e.g. Gaussian. $\theta$ uniquely defines a GAN. Unlike traditional GANs, we use a Bayesian approach and treat $\theta$ as random variables which conform to some prior distribution parameterized by $\Phi=[\Phi^g, \Phi^d]$. The inference of $\theta$ can be conducted by iteratively sampling~\cite{Saatchi_Bayesian_17}: 
\begin{align}
    p(\theta^g | &Z, \theta^d) \propto (\prod_{i=1}^{N^g} D(G(z_{(i)}; \theta^g);\theta^d))p(\theta^g|\Phi^g) \label{eq:Gsampling}\\
    p(\theta^d | &Z, X, \theta^g) \propto \prod_{i=1}^{N^d}D(x_{(i)};\theta^d) \times \nonumber \label{eq:Dsampling}\\ &\prod_{i=1}^{N^g}(1 - D(G(z_{(i)};\theta^g);\theta^d)) \times p (\theta^d|\Phi^d)
\end{align}
where $N^g$ and $N^d$ are the total numbers of generated and real images, $p(\theta^g|\Phi^g)$ and $p(\theta^d|\Phi^d)$ are \textit{prior} distributions of the network weights and sometimes combined as $p(\theta|\Phi)$ for simplicity. For our goals, the choice of the prior is based on the following consideration. First, if $X$ has $K$ modes corresponding to $K$ clusters, we aim to learn $K$ mappings through $K$ distinctive GANs, and each GAN is only responsible for generating one cluster of images. This dictates that the draws from the prior needs to be discrete. Second, since the $K$ value is unknown \textit{a priori}, we need to assume that $K\rightarrow\infty$. Therefore, we employ a Dirichlet Process (DP) as the prior $p(\theta|\Phi)$ for $\theta$s. 

A $DP(\alpha, \Phi)$ is a distribution of probabilistic distributions, where $\alpha$ is called \textit{concentration} and $\Phi$ is the \textit{base} distribution. It describes a `rich get richer' sampling~\cite{Neal_Markov_2000}:
\begin{equation}
\label{eq:CRP1}
    \theta_i | \mathbf{\theta_{-i}}, \alpha, \Phi \sim \sum_{l=1}^{i-1}\frac{1}{i-1+\alpha} \delta_{\theta_l} + \frac{\alpha}{i-1+\alpha}\Phi
\end{equation}
where an infinite sequence of random variables $\theta$s are i.i.d. according to $\Phi$. $\mathbf{\theta_{-i}}=\{\theta_1,\dots,\theta_{i-1}\}$. $\delta_{\theta_l}$ is a delta function at a previously drawn sample $\theta_l$. When a new $\theta_i$ is drawn, either a previously drawn sample is drawn again (with a probability proportional to $\frac{1}{i-1+\alpha}$), or a new sample is drawn (with a probability proportional to $\frac{\alpha}{i-1+\alpha}$). Assuming each $\theta$ has a value $\phi$, there can be multiple $\theta$s having the same value $\phi_k$. So there are only $K$ distinctive values in a total of $i$ samples drawn so far in Equation \ref{eq:CRP1} where $K < i$.  An intuitive (but not rigorous) analogy is rolling a dice multiple times. Each time one side (a sample) is chosen but overall there are only $K=6$ possible values. 

To see the 'rich get richer' property, the more $\phi$ has been drawn before, the more likely it will be drawn again. This property is highlighted by another equivalent representation called Chinese Restaurant Processes (CRP) \cite{Aldous_exchangeability_1983}, where the number of times the $k$th ($k\in K$) value $\phi_k$ has been drawn is associated with its probability of being drawn again:

\begin{equation}
\label{eq:CRP2}
    \theta_i | \mathbf{\theta_{-i}}, \alpha, \Phi \sim \sum_{k=1}^{K}\frac{N_k}{i-1+\alpha} \delta_{\phi_k} + \frac{\alpha}{i-1+\alpha}\Phi
\end{equation}
where $\delta_{\phi_k}$ is a delta function at $\phi_k$, and $N_k$ is how many times that $\phi_k$ has been sampled so far. Equation~\ref{eq:CRP1} and~\ref{eq:CRP2} are equivalent with the former represented by draws and the latter by actual values. 

\subsection{Mixture of Infinite GANs}
We propose a new Bayesian GAN which is a mixture of infinite GANs model. Following Eq.~\ref{eq:CRP2}, $\phi$ represents the network weights of a GAN. Imagine we have $K\rightarrow \infty$ GANs and examine $X$ one by one. For each image $x_i$, we sample the best GAN $\phi_{c_i}$ (based on some criteria) to generate it. So $N_k = \sum \mathbf{1}_{c_i = k}$ is the total number of images already selecting the $k$th GAN. The more frequently a GAN is selected, the more likely it will be selected in future. If all $N_k$s are small, then a new GAN is likely to be sampled based on $\Phi$. We  describe the generative process of our model as:
\begin{align}
\label{eq:generation}
&\text{sample } z_i \sim Z, \{\phi_1,\dots,\phi_k,\dots,\phi_K\}\sim\Phi \nonumber \\
&\text{sample } c_i \sim CRP (\alpha, \Phi; c_1, \dots, c_{i-1}), \text{where $c_i=k$} \nonumber \\
&x_i = G(z_i; \phi^g_k) \text{ so that } D(x_i;\phi^d_k) = 1
\end{align}
where $c_i$ now is an indicator variable, $\phi_{\{c_i=k\}}$=[$\phi^g_k$, $\phi^d_k$] are the parameters of the $k$th GAN. Combining Equation \ref{eq:CRP2}-\ref{eq:generation} with~\ref{eq:Gsampling}-\ref{eq:Dsampling}, the inference of our new model becomes:
\begin{align}
    p(\phi|\Phi&) = p(c_i|\mathbf{c_{-i}}) \sim CRP(\alpha, \Phi, \mathbf{c_{-i}}) \ \ c\in[1, K] \label{eq:Ksampling}\\
    p(\phi_k^g | &Z, \phi^d_k) \propto (\prod_{i=1}^{N^g_k} D(G(z_{(i)}; \phi^g_k);\phi^d_k))p(\phi^g_k|\Phi^g) \label{eq:Gsampling2}\\
    p(\phi^d_k | &Z, X, \phi^g_k) \propto \prod_{i=1}^{N^d_k}D(x_{(i)};\phi^d_k) \times \prod_{i=1}^{N^g_k}(1 - \nonumber \label{eq:Dsampling2}\\ D(G&(z_{(i)};\phi^g_k);\phi^d_k)) \times p (\phi^d_k|\Phi_d), \ 1 \leq k \leq K
\end{align}
where $\mathbf{c_{-i}} = \{c_1,\dots,c_{i-1}\}$. Sampling $c$ in Equation \ref{eq:Ksampling} will naturally compute the right value for $K$, essentially conducting unsupervised clustering~\cite{Neal_Markov_2000}.

\textbf{Classical GANs as maximum likelihood}. Equation~\ref{eq:Ksampling}-\ref{eq:Dsampling2} is a Bayesian generalization of classic GANs. If a uniform prior is used for $\Phi$ and iterative maximum a posteriori (MAP) optimization is employed instead of sampling the posterior, then the local minima give the standard GANs~\cite{goodfellow2017nips}. However, even with a flat prior, there is a big difference between Bayesian marginalization over the whole posterior and approximating it with a point mass in MAP~ \cite{Saatchi_Bayesian_17}.  Equation~\ref{eq:Ksampling}-\ref{eq:Dsampling2} is a specification of Equation~\ref{eq:Gsampling}-\ref{eq:Dsampling} with a CRP prior. Although Bayesian generalization over GANs have been explored before, we believe that this is first time a family of non-parametric Bayesian priors have been employed in modeling the distribution of GANs. Further, MIC-GANs aim to capture one cluster of images with one GAN in an unsupervised manner. The CRP prior can automatically infer the right value for $K$ instead of pre-defining one as in existing approaches \cite{liu_self_2020,mukherjee2019clustergan,deliGAN17}, where overestimating $K$ will divide a cluster arbitrarily into several ones while underestimating $K$ will mix multiple clusters.

\subsection{Mixture of Infinite Conditional GANs}
Iterative sampling over Equation \ref{eq:Ksampling}-\ref{eq:Dsampling2} would theoretically infer the right values for $\phi$s and $K$. While $\phi^d$ and $\phi^g$ can be sampled~\cite{Saatchi_Bayesian_17} or approximated~\cite{goodfellow2017nips}, $K$ needs to be sampled indirectly by sampling $c$, which turns out to be extremely challenging. To see the challenges, we need to analyze the full conditional distribution of $c$. To derive the full distribution, we first give the conditional probability of $c$ based on Eq.~\ref{eq:CRP2} as in~\cite{Neal_Markov_2000}: 
\begin{align}
\label{eq:CRPIndicator}
    p(c_i = k | \mathbf{c_{-i}}) \propto \frac{N_k}{i-1+\alpha} \nonumber \\
    p(c_i \ne c_j \text{ for all }j < i  | \mathbf{c_{-i}}) \propto \frac{\alpha}{i-1+\alpha}
\end{align}
Given the likelihood $p(x_i|z_i, \phi_k)$, or $p(x_i|\phi_k)$ if we omit $z_i$ as it is independently sampled, we now combine it with Equation \ref{eq:CRPIndicator} to obtain the full conditional distribution of $c_i$ given all other indicators and the current $x_i$:
\begin{align}
\label{eq:CRPFull}
    p(c_i=k | \mathbf{c_{-i}}, x_i, \mathbf{\phi}) \propto &\frac{N_k}{i - 1 + \alpha}p(x_i|\phi_k) \text{ if } \phi_k \text{ exists,    } \nonumber \\
    p(c_i = c_{new} | \mathbf{c_{-i}}, x_i, \mathbf{\phi}) \propto &\frac{\alpha}{i - 1 + \alpha} \int p(x_i|\phi)d\Phi \nonumber \\
    &\text{if a new $\phi_{new}$ is needed}
\end{align}
where if a $\phi_{new}$ is needed then it will be sampled from the posterior $p(\phi|x_i, \Phi)$. Eq.~\ref{eq:CRPFull} is used to sample Eq.~\ref{eq:Ksampling}.

\subsubsection{Challenges of Inference} 
One method to sample $c_i$ is Gibbs sampling \cite{Neal_Markov_2000}, which requires the prior $p(\phi|\Phi)$ to be a conjugate prior for the likelihood; otherwise additional sampling (e.g. Metropolis-Hasting) is needed to approximate the integral in Equation \ref{eq:CRPFull}. However, for MIC-GANs, not only is the prior not a conjugate prior for the likelihood, neither the likelihood nor the posterior can be even explicitly represented, which bring the following challenges: (1) Unable to directly compute the likelihood $p(x_i|\phi_c)$, which is a well-known issue for GANs \cite{Eghbalzadeh2017Likelihood}. (2) The sampling from $p(\phi|x_i, \Phi)$ is ill-conditioned. Since the prior $\Phi$ cannot be explicitly represented, direct sampling from $p(\phi|x_i, \Phi)$ becomes impossible. Alternatively, methods such as Markov Chain Monte Carlo are theoretically possible. But the dimension is often high, which will make the sampling prohibitively slow. (3) Sampling will dynamically change $K$. Each time $K$ grows/shrinks, a GAN needs to be created/destroyed, which is far from ideal in terms of the training speed. This is also an issue of existing methods with multiple GANs~\cite{lee2020neural,hoang_mgan_2018}. 

\subsubsection{Enhanced Model for Inference} 

To tackle challenge (2), we introduce a conditional variable $C$ while forcing all GANs to share the same $\phi$ so that they become $G_{\phi^g}(X|Z,C_k)$ and $D_{\phi^d}([0,1]|X, C_k)$ instead of $G(X|Z;\phi^g_k)$ and $D([0,1]|X, \phi^d_k)$ respectively, where $C\sim p(C)$ is well-behaved, e.g. a multivariate Gaussian. This formulation is similar to Conditional GANs (CGANs) but with a Bayesian treatment on $C$. Indeed, by introducing a conditional variable into multiple layers in the network, we exploit its ability of changing the mapping. Also, we now only need one GAN parameterized by $\phi=[\phi^g, \phi^d]$ and eliminate the need for multiple GANs without compromising the ability of learning $K$ distinctive mappings. Now the role of $C$ is the same as $\Phi$ in Equation \ref{eq:generation}, leading to:

\begin{align}
\label{eq:generation2}
&\text{sample } z_i \sim Z, \{C_0,\dots,C_k,\dots,C_K\} \sim p(C) \nonumber \\
&\text{sample } c_i \sim CRP (\alpha, C; c_1, \dots, c_{i-1}) \text{, where $c_i=k$} \nonumber \\
&x_i = G_{\phi^g}(z_i, C_k) \text{ so that } D_{\phi^d}(x_i, C_k) = 1,
\end{align}
where sampling from the posterior $p(\phi|x_i,C)$ (previously $p(\phi|x_i,\Phi)$) in Equation~\ref{eq:CRPFull} becomes feasible. Note our formulation is different from traditional CGANs and GANs with multiple generators~\cite{hoang_mgan_2018,Sage_Logo_2018,Kundu_GANtree_2019,liu_self_2020} in: (1) our approach is still Bayesian. (2) we still model an infinite number of GANs and do not rely or impose assumptions on the prior knowledge of cluster numbers. (3) we do not need to actually instantiate multiple GANs.

Next, we still need to be able to compute the likelihood $p(x_i|\phi_k)$ in Equation~\ref{eq:CRPFull}, which is challenge (1). Now $p(x_i|\phi_k)$ becomes $p(x_i|C_k)$. Since GANs cannot directly compute likelihoods, we employ a surrogate model that can compute likelihoods while mimicking GAN's mapping behavior. Each $C$ corresponds to one cluster, so the GANs can be seen as a mapping between $C$s and images. This correspondence is exactly the same as classifiers. So we design a classifier $Q$ to learn the mapping so that $p(x_i|C_k) \propto Q(c=k | x_i)$, where $c$ is the same $c$ as in Equation~\ref{eq:CRPFull} but here is also a cluster label out of $K$ clusters. Existing image classifiers can approximate likelihoods, e.g. through softmax. However, our experiments show that softmax-based likelihood approximation tends to behave like discriminative classifiers which focus on learning the classification boundaries. In Equation \ref{eq:CRPFull}, we need a classifier with density estimation. We thus define $Q$ as:
\begin{align}
\label{eq:classifier}
    Q&(c_i = k | x_i,\phi^q) \propto\ p(x_i|C_k,\phi^q)p(C_k|C)p(\phi^q|\Phi) \nonumber \\
    &= \mathcal{N}(x_i | \mu_k, \Sigma_k,\phi^q)p(C_k|C)p(\phi^q|\Phi) \nonumber \\
    &=\mathcal{N}(y_i | \mu_k, \Sigma_k)p(y_i|x_i,\phi^q)p(C_k|C)p(\phi^q|\Phi)
\end{align}
where $\mathcal{N}$ is a Normal distribution. $\mu_k$ and $\Sigma_k$ are the mean and covariance matrix of the $k$th Normal. $Q$ is realized by a deep neural net, parameterized by $\phi^q$ with a prior $p(\phi^q|\Phi)$, and classifies the latent code $y_i$ of $x_i$ by a Infinite Gaussian Mixture Model (IGMM)~\cite{Rasmussen_infinite_1999}.  To see this is an IGMM, $p(C_k|C)$ is the same CRP as in Equation~\ref{eq:generation2}, so that now the conditional variable $C$ and the Gaussian component $[\mu, \Sigma]$ are coupled (through indices $c$ under the same CRP prior). The inference on the Bayesian classifier $Q$ can be done through iteratively sampling:

\begin{align}
    p(\phi^q|X, \mu, \Sigma) \propto& \prod^N_{i=1}( \sum^K_{k=1} \mathcal{N}(y_i | \mu_k, \Sigma_k)p(y_i|x_i, \phi^q)) \nonumber \\ \times &p(\phi^q|\Phi^q) \label{eq:QSampling} \\
    p(\mu_{c_i}, \Sigma_{c_i} | x_i, c_i =&\ k, \mathbf{x_{-i}}, \mathbf{c_{-i}}, \phi^q) = \int\int p(y_i|\mu_k,\Sigma_k) \nonumber \label{eq:IGMMSampling}\\
    p(y_i|x_i, &\phi^q)p(\mu_k, \Sigma_k | \mathbf{x_{-i}}, \mathbf{c_{-i}}) \partial\mu_k\partial\Sigma_k
\end{align}
where $N$ is total number of images, $K$ is total number of clusters, $\mathbf{x_{-i}}$ is all images except $x_i$ with $\mathbf{c_{-i}}$ as their corresponding indicators. A Normal-Inverse-Warshart prior can be imposed on $[\mu, \Sigma]$ and we refer the readers to~\cite{Rasmussen_infinite_1999} for details. Alternatively, we could use IGMM directly on the images, but better classification results are achieved by classifying their latent features, as in the standard setting of many deep neural network based classifiers. We realize $p(y_i|x_i, \phi^q)$ as an encoder in $Q$.

Lastly for challenge (3), to avoid dynamically growing and shrinking $K$, we employ a truncated formulation inspired by \cite{Ishwaran_Gibbs_2001}, where an appropriate truncation level can be estimated. Essentially, the formulation requires us to start with a sufficiently large $K$ then learn how many modes are actually needed, which is automatically computed due to the aggregation property of DPs. Note that the truncation is only for the inference purpose and does not affect the capacity of MIC-GAN to model an infinite number of GANs. We refer the readers to \cite{Ishwaran_Gibbs_2001} for the mathematical proofs. 

\subsection{Adversarial Chinese Restaurant Process}
Finally, we have our full MIC-GANs model (Eq.~\ref{eq:generation2}-\ref{eq:classifier}) and are ready to derive our new sampling strategy called Adversarial Chinese Restaurant Process (ACRP). A Bayesian inference can be done through Gibbs sampling: iteratively sampling on $\phi^g$, $\phi^d$, $c$ (Equation~\ref{eq:Ksampling}-\ref{eq:Dsampling2},\ref{eq:CRPFull}), $\mu$, $\Sigma$  and $\phi^q$ (Equation~\ref{eq:classifier}-\ref{eq:IGMMSampling}). However, this tends to be slow given the high dimensionality. We thus propose to combine two schemes: optimization based on stochastic gradient descent and Gibbs sampling. While the former is suitable for finding the local minima that are equivalent to using flat priors and MAP optimization on $[\phi^g, \phi^d, \phi^q]$ \cite{Saatchi_Bayesian_17}, the latter is suitable for finding the posterior distributions of $c$, $\mu$ and $\Sigma$. We give the general sampling process of ACRP in Alg.\ref{alg:ACRP} and refer the readers to the supplementary materials for details.

In non-parametric Bayesian, $p(C)$ is governed by hyper-parameters which can be incorporated into the sampling. However, unlike traditional non-parametric Bayesian where a $C$ would specify a generation process, our generation process is mainly learned through the GAN training. In other words, the shape of $p(C)$ is not essential, as long as $C$s are distinctive, i.e. conditioning different mappings. So we fix $p(C)$ to be a multivariate Gaussian without compromising the modeling power of MIC-GANs.

Finally, since the learned GANs are governed by a DP, another equivalent representation is $G(X|Z,C) = \sum_{k=1}^{\infty}\beta_k G_k(X | Z, C_k)$, where the subscript $k$ indicates the $k$th GAN. $\beta_k$ is a weight and $\sum_{k=1}^{\infty}\beta_k = 1$. This is another interpretation of DP called \textit{stick-breaking} where $\beta_k = v_k\prod_{j=1}^{k-1}v_j$ and $v \sim Beta$ \cite{Sethuraman_constructive_1994}. After learning, the weights $\beta$s can be computed by the percentage of images assigned to each GAN. The stick-breaking interpretation indicates that the learned $G$s form a \textit{convex combination} of clusters, which is an active construction of the latent space. This enables controlled generation e.g. using single $C$s to generate images in different clusters, and easy exploration of the data space e.g. via interpolating $C$s.

\begin{algorithm} 
\caption{Adversarial Chinese Restaurant Process} 
\label{alg:ACRP} 
\begin{algorithmic}
\Require  ~~\\
$epochs$ - the number of total training epochs; \\
N is the total number of images; \\
Initialize all variables (supplementary material);
\end{algorithmic} 
\begin{algorithmic}[1] 
\For{$epoch=1$ to $epochs$}

\State For $x_i$ in $\mathcal{X}$, classify $x_i$ via Eq.~\ref{eq:classifier};
\State Compute $\{N_k\}_{k=1}^K$ in Eq.~\ref{eq:CRPFull};
\State For {$k=1$ to $K$} Sample $\mu_k$ and $\Sigma_k$ via Eq.~\ref{eq:IGMMSampling};
\State Sample $\{c_i\}_{i=1}^N$ via Eq.~\ref{eq:CRPFull};
\State Optimize $\phi^g$ and $\phi^d$ via conditional GAN loss;
\State Optimize $\phi^q$ via Maximum Likelihood (Eq.~\ref{eq:QSampling});
\EndFor
\end{algorithmic} 
\end{algorithm}

\textbf{Implementation details}. Due to the space limit, please refer to the supplementary materials for the details of models, data processing, training settings and performances.


\section{Experiments}


We adopt StyleGAN2~\cite{styleGAN2}, StyleGAN2-Ada~\cite{Karras2020ada} and DCGAN~\cite{radford2015unsupervised} to validate that MIC-GANs can incorporate different GANs. We employ several datasets for extensive evaluation, including MNIST~\cite{mnist}, FashionMNIST~\cite{xiao2017online}, CatDog from~\cite{choi2020stargan} and CIFAR-10~\cite{cifar10}. Moreover, we build a challenging dataset, named Hybrid, consisting of data with distinctive distributions. It is a mixture of the `0', `2', `4', `6', `8'  from MNIST, the `T-shirt', `pullover', `trouser', `bag', `sneaker'  from FashionMNIST, the cat images from Catdog and human face images from CelebA~\cite{liu2015faceattributes}.






 For comparison, we employ as baselines several state-of-the-art GANs including DMGAN~\cite{khayatkhoei2018disconnected}, InfoGAN~\cite{infoGAN16}, ClusterGAN~\cite{mukherjee2019clustergan}, DeliGAN~\cite{deliGAN17}, Self-Conditioned GAN~\cite{liu_self_2020} and StyleGAN2-Ada~\cite{Karras2020ada}, whose code is shared by the authors. Notably, DeliGAN, InfoGAN, ClusterGAN and Self-Conditioned GAN need a pre-defined cluster number, while MIC-GANs and DMGAN compute it automatically. To be harsh on MIC-GANs, we provide a large cluster number as an initial guess to MIC-GANs and DMGAN, while giving the ground-truth class number to the other methods.

\subsection{Unsupervised Clustering}
Although MIC-GANs focus on unsupervised image generation, it clusters data during learning. We evaluate its clustering ability by Cluster Purity~\cite{ClusterPurity08} which is a common metric to measure the extent to which clusters contain a single class: $Purity = \frac{1}{N}\sum_{i=1}^K{max_j|c_i \cap t_j|}$, where $N$ is the number of images, $K$ is the number of clusters, $c_i$ is a cluster and $t_j$ is the class which has the highest number of images clustered into $c_i$. The purity reaches the maximum value 1 when every image is clustered into its own class. Since MIC-GANs compute $K$ clusters, we rank them by their $\beta_k$s in the descending order and choose the top $n$ clusters for purity calculation, where $n$ is the ground-truth class number. Note the ground-truth is only used in the testing phase, not in the training phase.


\begin{table}[tb]
  \begin{center}
  \begin{tabular}{|c|c|c|c|c|}
  \hline
   & MNIST & FashionMNIST & CatDog & Hybrid \\
  \hline
  Purity & 0.9489 & 0.6362 & 0.9736 & 0.9567 \\
  \hline
  FID & 12.79 & 17.97 & 26.24 & 11.2 \\
  \hline
  \end{tabular}
  \caption{Purity and FID scores from MIC-GANs.}
  \vspace{-0.3cm}
  \label{tab:purity_fid}
  \end{center}
\end{table}

We use $K$=20, 20, 4 and 25 for MINST, FashionMNIST, CatDog and Hybrid. The results are shown in Table \ref{tab:purity_fid}. MIC-GANs achieve high purity on MNIST, CatDog and Hybrid. On FashionMNIST, the purity is relatively low. We find that it is mainly due to the ambiguity in the class labels. One example is that the bags with/without handles are given the same class label in FashionMNIST but divided by MIC-GANs into two clusters, as shown later. This finer-level clustering is automatically conducted, which we argue is still reasonable albeit leading to a lower purity score.


\begin{table}[tb]
  \begin{center}
  \begin{tabular}{|c|c|c|c|c|c|c|}
  \hline
  & \multicolumn{3}{c|}{MNIST} & \multicolumn{3}{c|}{Hybrid}\\
  \hline
   & K & Purity & FID & K & Purity & FID\\
  \hline
  1 & 20 & \textbf{0.9489} & 12.79 & 25 & \textbf{0.9567} & \textbf{11.2} \\
  2 & 20 & 0.7612 & \underline{10.82} & 25 & 0.8959 & 32.31 \\
  3 & 10 & 0.8589 & 11.9 & 12 & 0.881 & 71.84 \\
  4 & 10 & \underline{0.8784} & \textbf{9.27} & 12 & \underline{0.9457} & \underline{19.15} \\
  5 & 10 & n/a & 127.58 & 10 & n/a & 241.69 \\
  \hline
  \end{tabular}
  \caption{Purity and FID of MIC-GANs(1), DMGAN(2), InfoGAN(3), ClusterGAN(4) and DeliGAN(5).}
  \vspace{-0.7cm}
  \label{tab:acc_fid}
  \end{center}
\end{table}

As a comparison, we also evaluate the baseline methods on MNIST and Hybrid, with the ground-truth class number $K$=10 for InfoGAN and ClusterGAN, and $K$=20 for DMGAN. Note that DeliGAN does not have a clustering module, so we exclude it from this evaluation. For both MIC-GANs and DMGAN, we use the top $n$ clusters where $n$ is the ground-truth class number. As we can see in Table~\ref{tab:acc_fid}, surprisingly, MIC-GANs achieve the highest purity score ($0.9489$) among these methods, even without using the ground-truth class number in training. It demonstrates that the MIC-GANs can effectively learn the modes in the MNIST without any supervision. In addition, we also conduct the comparison on Hybrid, with the ground-truth $K$=12 for InfoGAN and ClusterGAN , and $K$=25 for DMGAN. As shown in Table~\ref{tab:acc_fid}, MIC-GANs achieve the best purity score ($0.9567$). ClusterGAN ($0.9457$) comes a close second. However, we provide the ground-truth class number to ClusterGAN which is strong prior knowledge, while MIC-GANs have no such information.

\subsection{Image Generation Quality}
\begin{figure*}[t]
  \centering
  \includegraphics[width=1.0\linewidth]{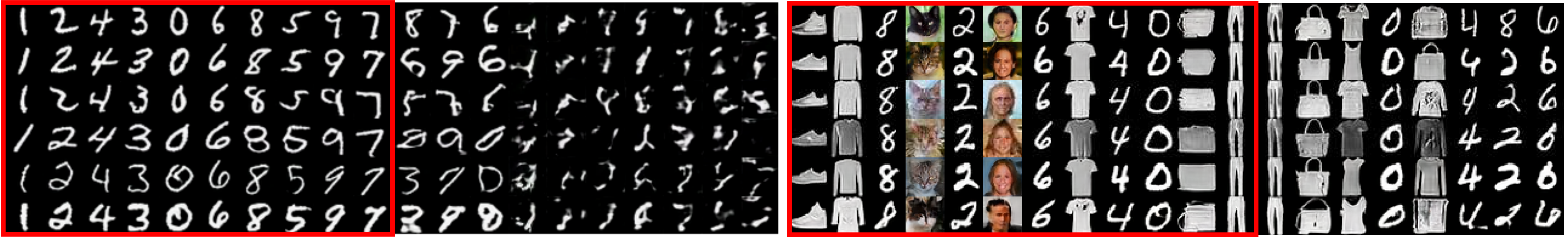}
  \caption{Our results on MNIST (left) and Hybrid (right) dataset, both with $K=25$. Each column is generated from a mode, and the columns are sorted by $\alpha$s (the last 4 modes are not shown). The red boxes mark the top $n$ modes in the results.}
  \vspace{-0.3cm}
\label{fig:ours_mnist_hybrid}
\end{figure*}

\begin{figure}[t]
  \centering
  \includegraphics[width=1.0\linewidth]{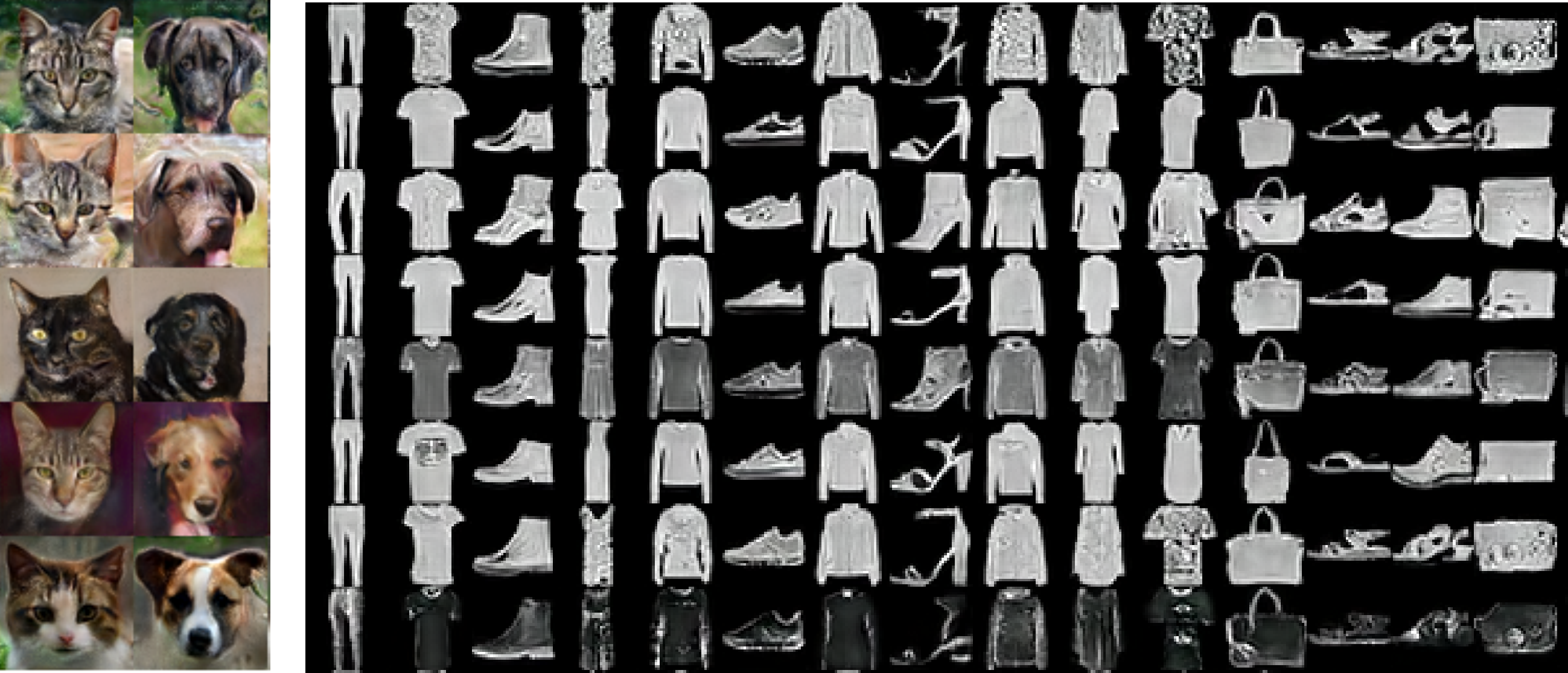}
  \caption{Our results on CatDog (left) and FashionMNIST (right) dataset. Each column is generated from one mode, and the columns are sorted by $\alpha$s. }
\vspace{-0.3cm}
\label{fig:ours_catdog_fashion}
\end{figure}
We conduct both qualitative and quantitative evaluations on the generation quality. For quantitative evaluations, we use Frechet Inception Distance (FID)~\cite{FID17} as the metric.  Qualitatively, generated samples can be found in Figure \ref{fig:ours_mnist_hybrid}-\ref{fig:ours_catdog_fashion}. Due to the space limit, we rank the modes based on their $\beta$s in the descending order and only show the top modes. More results can be found in the supplementary materials. Intuitively, all the modes are captured cleanly, shown by that the top modes contain all the classes in the datasets. This is where MIC-GANs capture most of the `mass' in the datasets. In addition, each mode fully captures the variation within the mode, no matter it is the writing style in MNIST, the shape in FashionMNIST or the facial features in CatDog and CelebA. The balance between clustering and within-cluster variation is automatically achieved by MIC-GANs in an unsupervised setting. This is very challenging because the within-cluster variations are distinctive in Hybrid since the data comes from four datasets. Beyond the top modes, the low-rank modes also capture information. But the information is less meaningful (the later modes in MNIST) or mixed (later modes in Hybrid) or contain subcategories such as the separation of bags with/without handles in FashionMNIST. However, this does not undermine MIC-GANs because the absolute majority of the data is captured in the top modes. Quantitatively, we found that good FID scores can be achieved, shown in Table \ref{tab:purity_fid}.

We further conduct comparisons on MNIST and Hybrid using the same settings as above for all methods. As demonstrated in Table~\ref{tab:acc_fid}, MIC-GANs obtain a comparable FID score to other methods without any supervision in MNIST. Besides DeliGAN, DMGAN and MIC-GANs achieve slightly worse FID scores. We speculate that this is because MIC-GANs and DMGANs do not use prior knowledge and therefore have a disadvantage. One exception is DeliGAN. In their paper, the authors chose a small dataset (500 images) for training and achieved good results. However, when we run their code on the full MNIST dataset, we were not able to reproduce comparable results even after trying our best to tune various parameters. Next, in the challenging Hybrid dataset whose distribution is more complex than MNIST, MIC-GANs achieve the best FID score. Without any supervision, MIC-GANs not only capture the multiple data modes well, but generate high-quality images.  We also conduct comparisons on CIFAR with Self-Conditioned GAN~\cite{liu_self_2020} and StyleGAN2-Ada~\cite{Karras2020ada}. To investigate how MIC-GANs compare with other methods on a dataset with different numbers of modes, we sample C=\{4,7,10\} classes from CIFAR where C=10 is the full dataset. We also adopt StyleGAN2-Ada in MIC-GANs for CIFAR. The results show that MIC-GANs can achieve better FID scores (Table~\ref{tab:comparison_cifar}). The change of baselines is mainly due to that we only compare MIC-GANs with methods on the datasets on which they are tested, for fairness.


\begin{figure*}[t]
  \centering
  \includegraphics[width=1.0\linewidth]{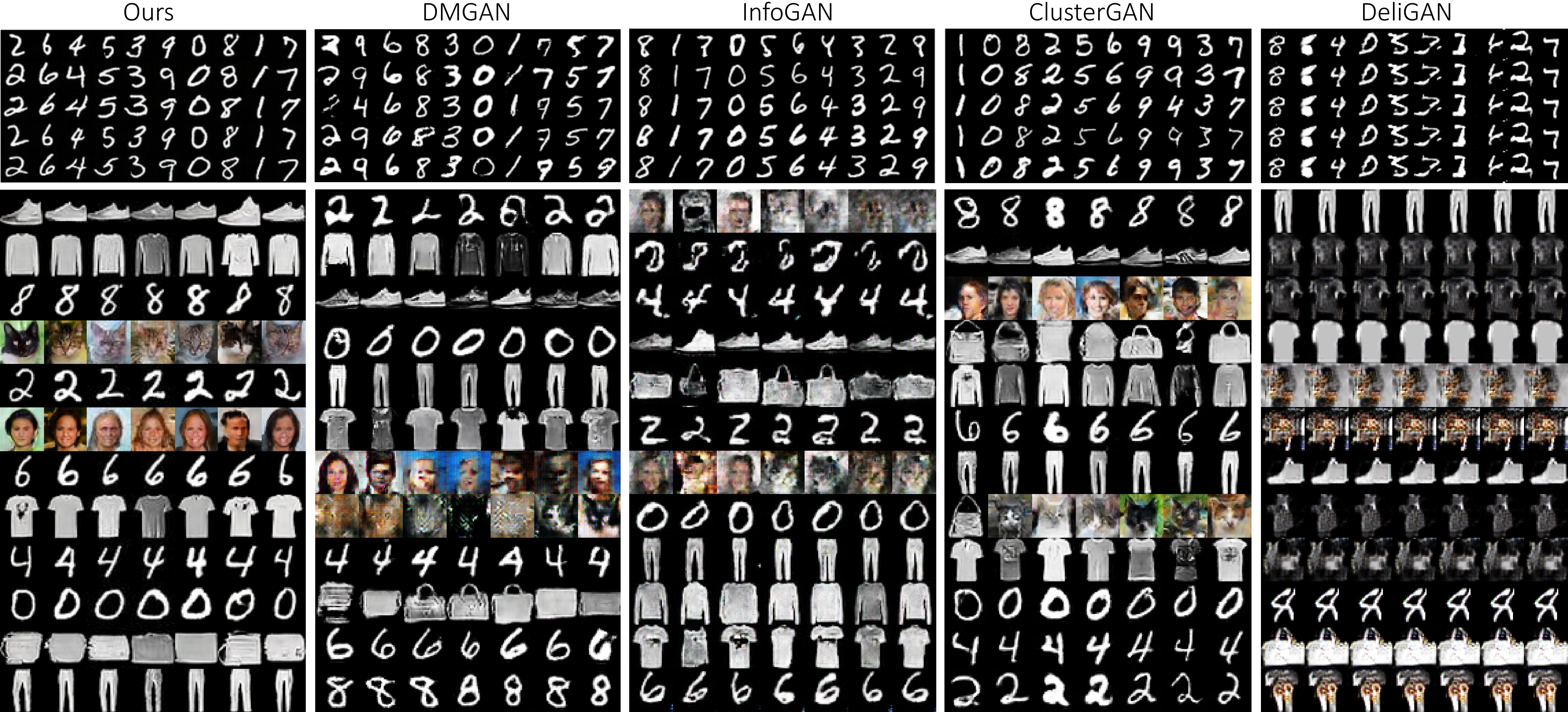}
  \caption{Generation comparisons on MNIST (top) and Hybrid (bottom) dataset. We use the ground-truth $K=10$ on MNIST and $K=12$ on Hybrid for InfoGAN, ClusterGAN and DeliGAN, and $K=20$ for MNIST and $K=25$ for Hybrid for MIC-GANs and DMGAN. Each column is generated from a mode for MNIST (top), and each row is generated from a mode for Hybrid (bottom).}
  \vspace{-0.3cm}
\label{fig:comparison_imgs}
\end{figure*}


\begin{table}
  \begin{center}
  \begin{tabular}{|c|c|c|c|c|c|c|c|c|}
  \hline
  \multicolumn{3}{|c|}{Ours} & \multicolumn{3}{c|}{SC-GAN} & \multicolumn{3}{c|}{StyleGAN2-Ada} \\
  \hline
  C & K & FID & C & K & FID & C & K & FID \\
  \hline
  4 & 10 & \textbf{5.31} & 4 & 10 & 237.96 & 4 & - & 5.64 \\
  \hline
  7 & 15 & \textbf{5.09} & 7 & 15 & 41.30 & 7 & - & 5.16 \\
  \hline
  10 & 20 & \textbf{4.72} & 10 & 20 & 22.00 & 10 & - & 5.03 \\
  \hline
  \end{tabular}
  \caption{Comparisons of our method, SC-GAN (Self-Conditioned GAN), and Styelgan2-Ada on CIFAR. `C' means the ground-truth class number in the dataset.}
  \vspace{-0.7cm}
  \label{tab:comparison_cifar}
  \end{center}
\end{table}

As a visual comparison, we show the generated images from different GANs in Figure~\ref{fig:comparison_imgs}. In the top row, MIC-GANs generate perceptually comparable results on MNIST to InfoGAN and ClusterGAN which were fed with the ground-truth class number, while achieving better clustering than DMGAN (e.g., mixed `9' and `4', `9' and `7') which is also unsupervised. In the challenging Hybrid dataset (bottom), MIC-GANs are able to generate high-quality images while correctly capturing all the modes. CIFAR images are shown in the supplementary materials.


\subsection{Ablation Study}
\begin{table}
  \begin{center}
  \begin{tabular}{|c|c|c|c|c|c|c|}
  \hline
  & \multicolumn{3}{c|}{DCGAN} & \multicolumn{3}{c|}{StyleGAN2} \\
  \hline
  \multirow{4}{*}{MNIST} & K & Purity & FID & K & Purity & FID \\
  \cline{2-7}
  & 1 & -      & 9.89 & 1  & -      & 12.96\\
  &15 & 0.9384 & 5.22 & 15 & 0.9397 & 9.92\\
  &20 & 0.9578 & 8.62 & 20 & 0.9489 & 12.79\\
  &25 & 0.93 & 9.13 & 25 & 0.9487 & 11.9 \\
  \hline
  \multirow{4}{*}{Hybrid} & K & Purity & FID & K & Purity & FID \\
  \cline{2-7}
  & 1 & -      & 60.83 & 1  & -     & 15.47\\  
  &15 & 0.9218 & 50.74 & 15 & 0.942 & 15.7\\
  &20 & 0.9611 & 48.17 & 20 & 0.923 & 13.31\\
  &25 & 0.966 & 45.07 & 25 & 0.9567 & 11.2 \\
  \hline
  \end{tabular}
  \caption{Purity and FID of ablation studies on MNIST and Hybrid.}
  \vspace{-0.7cm}
  \label{tab:ablation}
  \end{center}
\end{table}

  

We conduct ablation studies to test MIC-GANs' sensitivity to the free parameters. As an unsupervised and non-parametric method, there are not many tunable parameters which is another advantage of MIC-GANs. The main parameters are the $K$ value and the GAN architecture. We therefore test another popular GAN, DCGAN \cite{radford2015unsupervised} and vary the $K$ value. As shown in Table~\ref{tab:ablation}, the purity scores are very similar, which means the clustering is not significantly affected by the choices of GANs or the $K$ value. FID scores vary across datasets, which is mainly related to the specific choice of the GAN architecture. However, stable performance is obtained across different $K$s in every setting. In addition, the FID scores when $K=1$ are in general worse than those when $K>1$, confirming that our method can mitigate mode collapses. The same mode collapse mitigation can also be observed when comparing our method with StyleGAN2-Ada on CIFAR-10 (Table~\ref{tab:comparison_cifar}), where StyleGAN2-Ada is just our method with $K=1$. 


\subsection{Benefits of Non-parametric Learning}


\begin{figure}[t]
  \centering
  \includegraphics[width=0.99\linewidth]{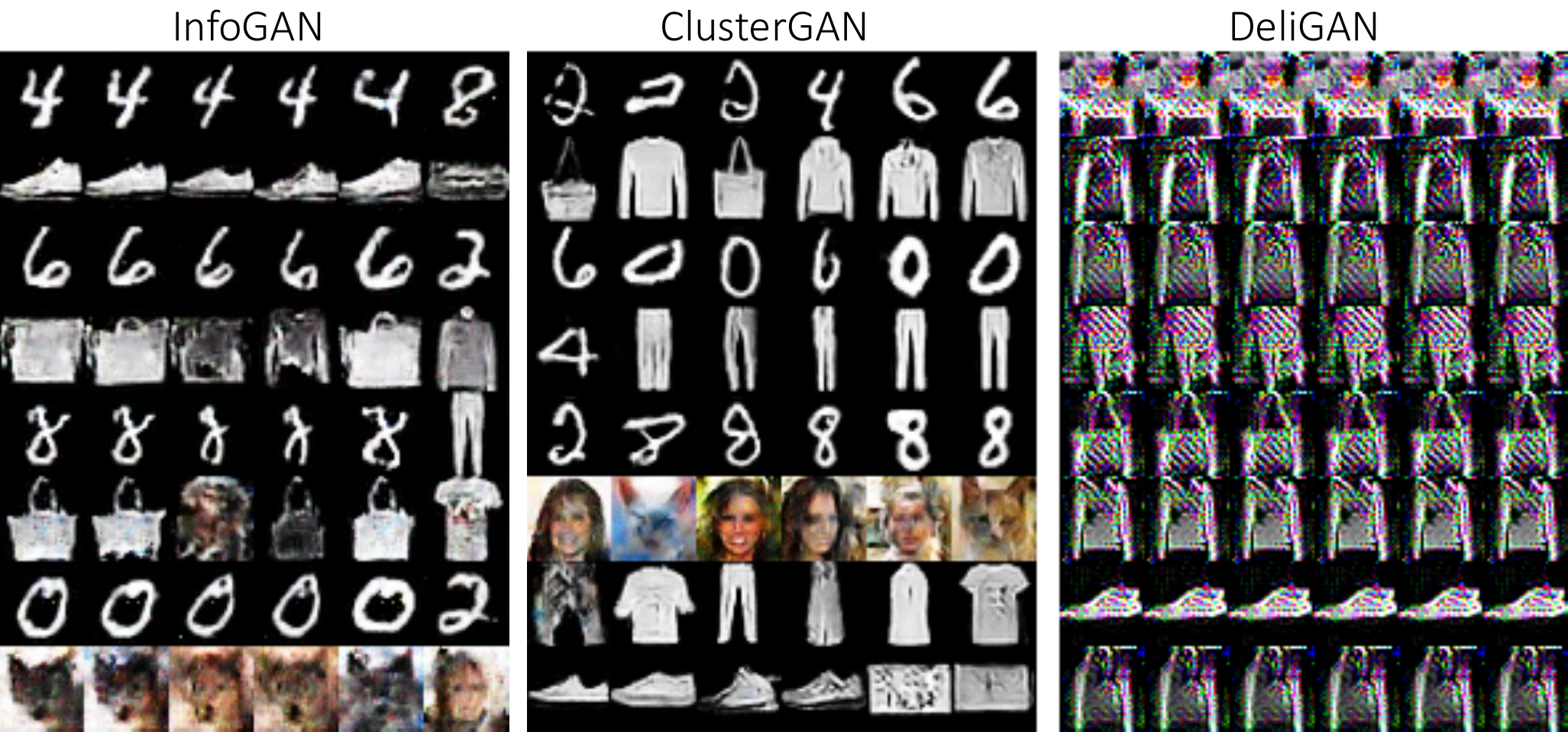}
  \caption{$K$ = 8 results of InfoGAN, ClusterGAN and DeliGAN.}
  \vspace{-0.3cm}
\label{fig:other_methods_imgs}
\end{figure}

In real-world scenarios, we do not often know the cluster number \textit{a priori}, under which we investigate the performance of InfoGAN, ClusterGAN and DeliGAN. We use Hybrid and run experiments with $K$=8, 12, 16 and 22 to cover $K$ values that are smaller, equal to and bigger than the ground-truth $K$=12. We only show the results of $K$=8 in Figure \ref{fig:other_methods_imgs} and refer the readers to the supplementary materials for fuller results and analysis. Intuitively, when $K$ is smaller than the ground-truth, the baseline methods either cannot capture all the modes or capture mixed modes; when $K$ is larger than the ground-truth, they capture either mixed modes or repetitive modes. In contrast, although MIC-GANs (Figure~\ref{fig:ours_mnist_hybrid}-\ref{fig:ours_catdog_fashion}) also learn extra modes, it concentrates the mass into the top modes resulting in clean and complete capture of modes. MIC-GANs are capable of high-quality image generation and accurate data mode capturing, while being robust to the initial guess of $K$.

\subsection{Latent Structure}
\begin{figure}[t]
  \centering
  \includegraphics[width=1.0\linewidth]{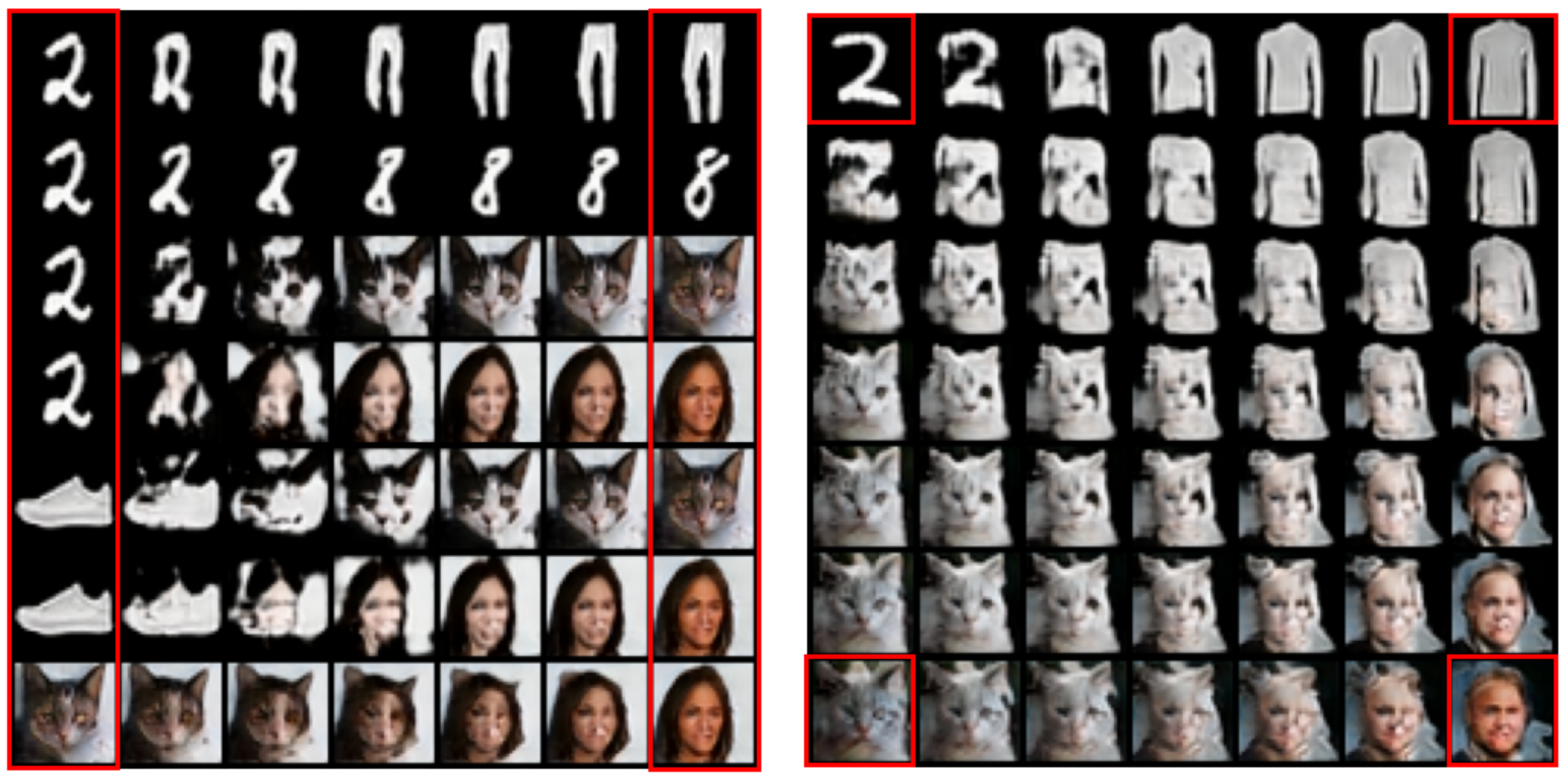}
  \caption{Left: each row is the interpolation results between two latent codes, where the first column and the last column are the original images. Right: the interpolation results among four latent codes, where each corner represents one mode.}
  \vspace{-0.3cm}
\label{fig:interpolation}
\end{figure}

Since MIC-GANs are a convex combination of GANs, we can do controlled generation, including using a specific mode, and interpolating between two or more different modes for image generation. Figure \ref{fig:ours_mnist_hybrid}-\ref{fig:ours_catdog_fashion} already show image generation based on single modes. We show interpolations between two $C$s and among four $C$s respectively in Figure \ref{fig:interpolation}. Through both bi-mode and multi-mode interpolation, we show that MIC-GANs structure the latent space well so that smooth interpolations can be conducted within the subspace bounded by the base modes. 


\section{Conclusion}
We proposed a new unsupervised and non-parametric generative framework MIC-GANs, to jointly tackle two fundamental GAN issues, mode collapse and unstructured latent space, based on parsimonious assumptions. Extensive evaluations and comparisons show that MIC-GANs outperform state-of-the-art methods on multiple datasets. MIC-GANs do not require strong prior knowledge, nor do they need much human intervention, providing a robust and adaptive solution for multi-modal image generation.
\\
\\
\textbf{Acknowledgments}\\
We thank anonymous reviewers for their valuable comments. The work was supported by NSF China (No. 61772462, No. 61890954, No. U1736217), the 100 Talents Program of Zhejiang University, and NSF under grants No. 2011471 and 2016414.
{\small
\bibliographystyle{ieee_fullname}
\bibliography{paper_arxiv}
}

\newpage
\begin{appendix}

\section{Algorithm Details}

The training of MIC-GANs is split into two stages, the initialization stage and the Adversarial Chinese Restaurant Process (ACRP) Stage. 

\subsection{Initialization Stage}

The initialization stage is to initialize the generator, enabling it to produce images of good quality. At the same time, we require the generator to produce conditioned output without supervised class labels.

The detailed algorithm for initialization is shown in Algorithm~\ref{Initial Training}. The training procedure is the same as the ordinary GANs, except that the generator is given a conditioned input. Note that the discriminator is not conditional, so it is not a conditional GAN. $Categ(\alpha_1, ..., \alpha_K)$ is the category distribution where index number $k$ is sampled according to the probability proportional to $\alpha_k$.The conditional input for generator is uniformly sampled, i.e. $\alpha_1, ..., \alpha_K=\frac{1}{K}$. 

After initialization, the generator can produce conditioned outputs. Generally, the outputs from one condition are more likely to be close to one class of images. However, because $K$ is different from the ground-truth class number, the outputs from one condition often either contain only a part of one class or multiple classes. Taking MNIST for example, when using $K=20$ which is larger than the ground-truth class number, after initialization there may be two modes generating '6' and one mode generating both '7' and '9'. However, it will be resolved in the later ACRP stage.

\subsection{ACRP Stage}

The main algorithm of ACRP is shown in Algorithm~\ref{alg:ACRP2}, and the Chinese Restaurant Process Sampling algorithm is shown in Algorithm~\ref{DP Sampling}. In practice, the encoder network in $Q$ is initialized before training in every epoch to prevent overfitting. The parameters of the GMMs need to be initialized before training. We let the covariance matrices $\Sigma_k$s be the identity matrix, and initialize the mean $\mu_k$s in the way that they become the vertices of a high-dimensional simplex and are equidistant to each other. This is to ensure that the Gaussians are distinctive. Next, the dimensionality of the latent space of the encoder in $Q$ needs to be decided. Theoretically, it can be any dimension that is smaller than that of the data space. In practice, we set the dimension of both the image embedding and GMM to $K$ which is the number of modes, so that conveniently the $\mu_k$s are the basis vectors of such a $K$-dimensional space. One straightforward solution is to use the one-hot $K$-vectors as the $\mu_k$s' initialization. As for the encoder loss $\mathcal{L}_Q(e, \mu_{c})$, we maximize the log likelihood of $e$ with respect to the Gaussian with $\mu_{c}$ as its mean.

\textbf{Likelihoods in GANs}. We do not solve the likelihood problem of GANs directly. MIC-GANs employ a surrogate density estimator. The design is due to the following reasons. First, MIC-GANs are designed to work essentially with any GANs. So the density estimation needs to be independent of the specific GAN architecture. Also, different GANs are designed for different tasks, e.g. StyleGAN, BigGAN, etc. They all contain specific architectures optimized for their aimed tasks. Therefore, we choose to keep any chosen GAN intact under MIC-GANs. Employing a surrogate density estimator is our current solution, and we are actively looking for a `true' solution. 

\begin{figure}[t]
  \centering
  \includegraphics[width=1.0\linewidth]{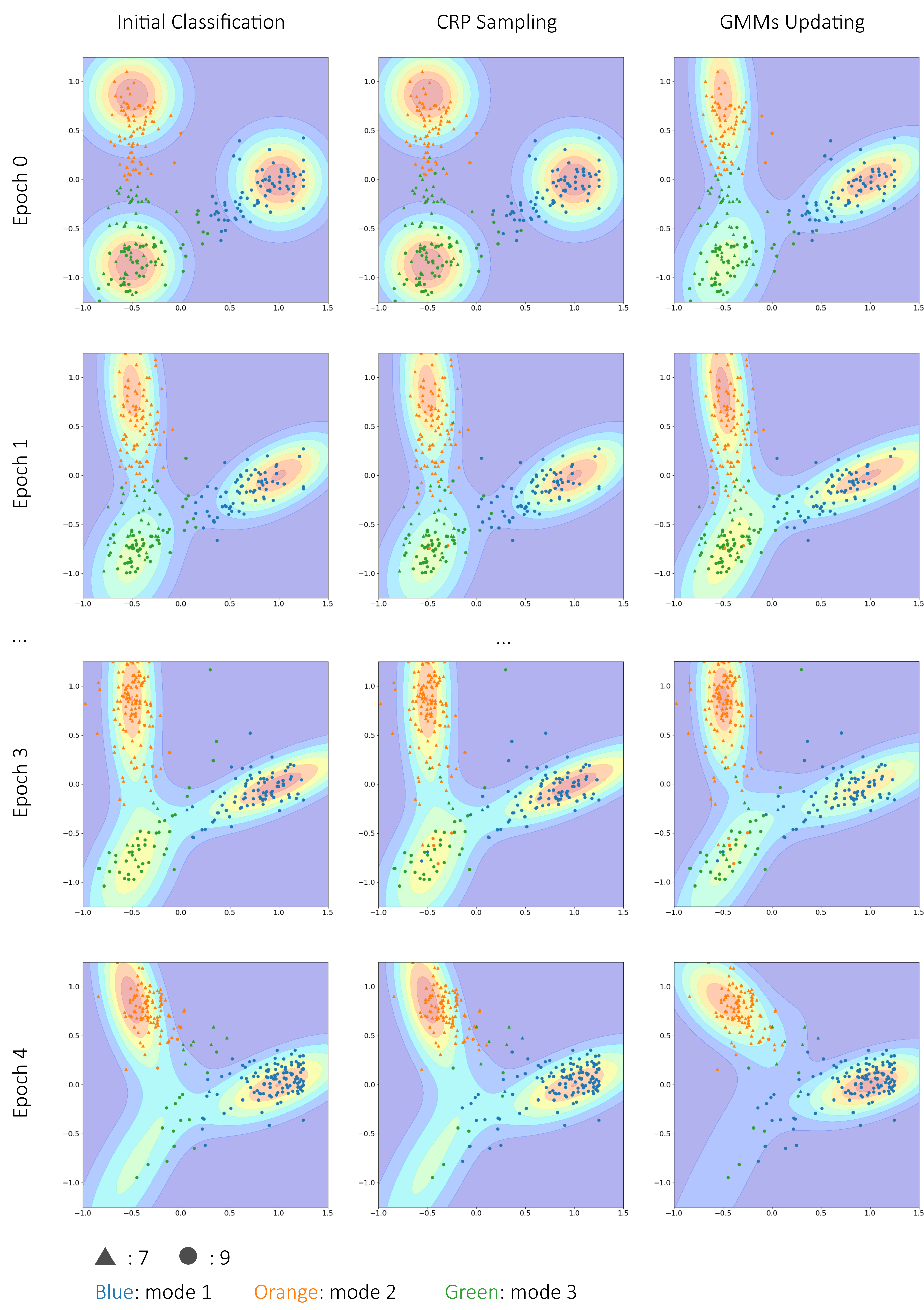}
  \caption{The transformation of classification results and GMMs in the ACRP stage. In each small figure, the small dots represent embeddings of training images from encoder $Q$, with the triangle dots representing number `7' and the circle dots representing number `9'. The colors of the dots represent the classifications to three modes. The background color visualizes the shape of GMMs. In each epoch, the left two figures show the classification results conducted directly from the gaussian probability and after CRP sampling and the right figure shows the updated GMMs based on the classification results after CRP sampling. }
\vspace{-0.3cm}
\label{fig:procedure}
\end{figure}

To help understand ACRP, we present a visualization (Figure~\ref{fig:procedure}) of the procedure of our algorithm with a simple dataset which only consists of number `7' and number `9' from MNIST. The figure visualizes line 7-18 of Algorithm~\ref{alg:ACRP2}. In the algorithm, we set the embedding dimension to be $2$ with $K=3, iters_1=1, iters_2=1$, and we fix the means of GMMs to be three vertices of an equilateral triangle for better visualization quality. As shown in Figure~\ref{fig:procedure}, at first, `mode 1' contains most `9's and `mode 2' contains most `7's while `mode 3' contains both `9's and `7's. With the algorithm progressing, the number of images classified to `mode 3' is gradually reduced, because more `9's that were originally classified to `mode 3' are now classified to `mode 1', and similarly `7's that are originally under `mode 3' are now classified to `mode 2'. In Epoch 4, most of the images are divided into two modes and the classification is almost correct. Meanwhile, the original `mode 3' basically disappears as the probability of it being sampled again becomes nearly zero. To see this, $N_3$ in line 12 in Algorithm~\ref{alg:ACRP2} after CRP becomes very small and the probability of `mode 3' being sampled again is proportional to $N_3$. Figure~\ref{fig:procedure} shows the `richer gets richer' property of Chinese Restaurant Process.

\begin{table}[tb]
    \centering
    \begin{tabular}{p{1cm}|c}
         Variable & Meaning \\
         \hline
          $\mathcal{X}$ & the whole real images \\
         \hline
         $K$ &  the number of modes \\
         \hline
         $N$ & the total number of real images \\
         \hline
         $G_{\phi^G}$ & generator parameterized by $\phi^G$ \\
         \hline
         $D_{\phi^D}$ & discriminator parameterized by $\phi^D$ \\
         \hline
         $Q_{\phi^Q}$ & classifier parameterized by $\phi^Q$ \\
         \hline
         $z$ & the input noise for generator \\
         \hline
         $\alpha_k$ & the sampling probability of mode $k$ \\
         \hline
         $c_i$ & the picked mode index for each real image $i$ \\
         \hline
         $\mu_k, \Sigma_k$ & the parameters of the $k$th Gaussian \\
         \hline
         $N_k$ & the number of real images associated with mode $k$ \\
         \hline
         $p_{i,k}$ & the likelihood of real image $i$ on mode $k$ \\
         \hline
         $e$ & the embedding of an image, from encoder Q \\
         \hline
    \end{tabular}
    \caption{Symbols of MIC-GANs Training}
    \vspace{-0.4cm}
    \label{tab:ACRP_symbols}
\end{table}

\begin{algorithm} 
\caption{Initialization} 
\label{Initial Training} 
\begin{algorithmic}
\Require  ~~\\
$epochs$ - the number of total training epochs; \\
$N_{init}$ - the number of images for initialization training;
\end{algorithmic} 
\begin{algorithmic}[1] 
\For{$n=1$ to $N_{init}$}
\State Sample $x \sim \mathcal{X}$
\State Sample $z \sim \mathcal{N}(0,1)$, $c \sim Categ(\alpha_1, ..., \alpha_K)$
\State Generate fake image $\hat{x} = G(z, c)$
\State Optimize $\phi^g$ and $\phi^d$ via GAN loss
\EndFor
\end{algorithmic} 
\end{algorithm}

\begin{algorithm} 
\caption{Adversarial Chinese Restaurant Process} 
\label{alg:ACRP2} 
\begin{algorithmic}
\Require  ~~\\
$epochs$ - the number of total training epochs; \\
$N_Q$ - the number of images for training encoder in each epoch; \\
$N_{GD}$ - the number of images for training generator and discriminator in each epoch; \\
$iters_1$ - the number of iterations for CRP sampling and GMM updating; \\
Initialize() ;
\end{algorithmic} 
\begin{algorithmic}[1] 
\For{$epoch=1$ to $epochs$}
\For{$n=1$ to $N_Q$} \algorithmiccomment{Train Q}
\State Sample $z \sim \mathcal{N}(0,1)$, $c \sim Categ(\alpha_1, ..., \alpha_K)$
\State Get embedding $e = Q(G(z, c))$
\State Optimize $\phi_Q$ via encoder loss $L_Q = \mathcal{L}_Q(e, \mu_{c})$
\EndFor
\For{$iter=1$ to $iters_1$} \algorithmiccomment{Classify x}
\For {$x_i$ in $\mathcal{X}$} \algorithmiccomment{Computing likelihood}
\State $e_i=Q(x_i)$
\State $p_{i,k} = Gauss(e_i|\mu_k, \Sigma_k)$ for $k=1$ to $K$
\EndFor
\State Sample $\{c_i\}_{i=1}^N$, $\{N_k\}_{k=1}^K$ via CRP (Alg.~\ref{DP Sampling})
\State $\mathbb{E}_k \gets \emptyset$ for $k=1$ to $K$
\State $\mathbb{E}_{c_i} \gets \mathbb{E}_{c_i} \bigcup \{e_i\}$ for each $e_i$
\For {$k=1$ to $K$} \algorithmiccomment{Update GMMs}
\State Update $\mu_k$ and $\Sigma_k$ with $\mathbb{E}_k$
\EndFor
\EndFor
\State $\alpha_k \gets \dfrac{N_k}{N}$
\For{$n=1$ to $N_{GD}$} \algorithmiccomment{Train GAN}
\State Sample $x_i \sim \mathcal{X}$, and fetch corresponding $c_i$
\State Sample $z \sim \mathcal{N}(0,1)$, $c \sim Categ(\alpha_1, ..., \alpha_K)$
\State Generate fake image $\hat{x} = G(z, c)$
\State Optimize $\phi^g$ and $\phi^d$ via conditional GAN loss
\EndFor
\EndFor
\end{algorithmic} 
\end{algorithm}

\begin{algorithm} 
\caption{Chinese Restaurant Process Sampling} 
\label{DP Sampling} 
\begin{algorithmic}
\Require  ~~\\
$iters_2$ - the number of iterations for CRP sampling; 
\end{algorithmic} 
\begin{algorithmic}[1] 
\State $N_k \gets 0$ for $k=1$ to $K$
\For {$x_i$ in $\mathcal{X}$} 
\State $c_i \gets argmax(\{p_{i,k}\}_{k=1}^K)$
\State $N_{c_i} = N_{c_i} + 1$
\EndFor
\For{$iter=1$ to $iters_2$} 
\For {$x_i$ in $\mathcal{X}$} 
\State $N_{c_i} = N_{c_i} - 1$
\State $\beta_k \gets N_k \cdot p_{i,k}$ for $k=1$ to $K$
\State $\beta_k \gets \dfrac{\beta_k}{\sum{\beta_k}} $ for $k=1$ to $K$
\State Sample $c_i \sim Categ(\beta_1, ..., \beta_K)$
\State $N_{c_i} = N_{c_i} + 1$
\EndFor
\EndFor
\end{algorithmic} 
\end{algorithm}

\section{Implementation Details}

\subsection{Network Architecture}

\begin{figure}[t]
  \centering
  \includegraphics[width=1.0\linewidth]{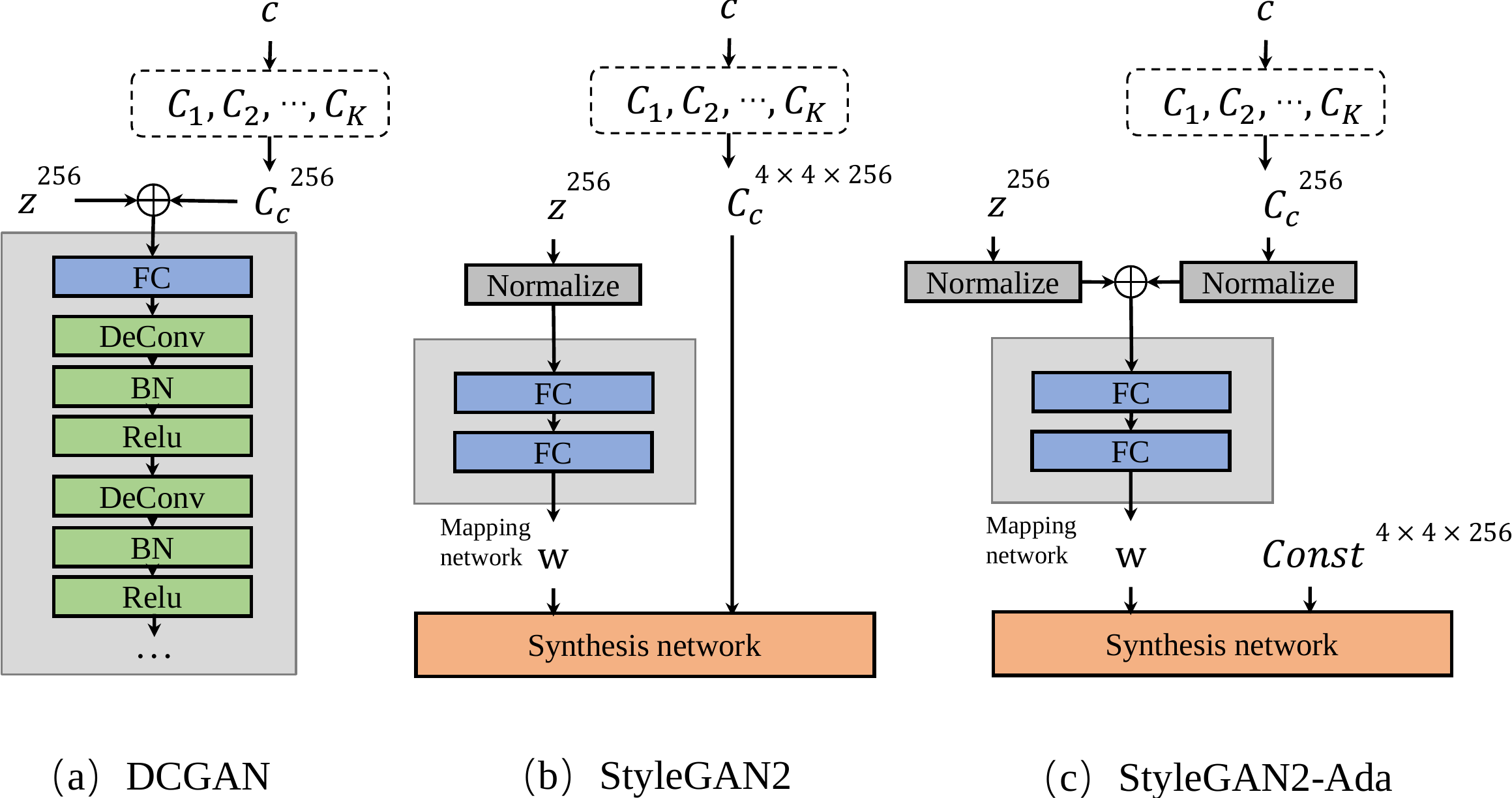}
  \caption{The conditional input heads of the generators for DCGAN, StyleGAN2 and StyleGAN2-Ada. }
\vspace{-0.3cm}
\label{fig:nets}
\end{figure}

We adopt different GAN models including DCGAN~\cite{radford2015unsupervised}, StyleGAN2~\cite{styleGAN2} and StyleGAN2-Ada~\cite{Karras2020ada} to validate our algorithm.
Specifically, in order to achieve conditioned generation, we modify the input of the generators to take conditions. The detailed implementation of the conditional inputs is shown in Figure~\ref{fig:nets}. For DCGAN and StyleGAN2-Ada, the condition of the generator is specified by adding the conditioned latent code $C_c$ to the noise $z$. In StyleGAN2, we tried to control the condition by picking one of $K$ constant inputs for the synthesis network. 

The network architecture of discriminator needs to be handled differently in different stages because the discriminator needs to take conditions in ACRP stage but the conditions are not reliable in the initialization stage. So we keep the discriminator as the original one in DCGAN or StyleGAN2 during initialization, and in ACRP stage, modify it to take conditions. For the discriminator of DCGAN and StyleGAN2, we follow the approach of traditional cGAN~\cite{cGAN14}, and for the discriminator of StyleGAN2-Ada, we follow the approach in~\cite{cGANProject18}. After initialization and at the beginning of the ACRP stage, a condition input is added to the discriminator.

For the network architecture of encoder in $Q$, we simply adopt a multi-layer convolutional network. For the dataset of MNIST, FashionMNIST, and Hybrid, we use a 4-layer CNN with a fully connected output layer, and for the dataset of CatDog, CIFAR and Tiny Imagenet, we use a 7-layer CNN with a fully connected output layer. Both of them do not use BatchNorm Layer.

Besides the DCGAN and StyleGAN, there are other GANs that are also suitable for mode separation, e.g., FiLM~\cite{perez2018film}.In fact, our algorithm can be applied to any conditional GANs theoretically.

\subsection{Training Details}

Images in MNIST, FashionMNIST, Hybrid, CIFAR and Tiny Imagenet are resized to $32$, and images in CatDog are resized to $64$.
When training, the batch size is set to $64$ for CatDog and CIFAR, and $256$ for the other datasets. 
During initialization, $N_{init}$ is set to $2400k$ for the MNIST FashionMNIST, Hybrid dataset, $1000k$ for the CatDog dataset, $2000k$ for the CIFAR and Tiny Imagenet dataset.
In ACRP stage, $N_Q$ is set to $64k$, and $N_{GD}$ is set to $300k$ for the MNIST FashionMNIST, Hybrid dataset, $100k$ for the CatDog dataset, $200k$ for the CIFAR and Tiny Imagenet dataset.
We trained the MIC-GANs for totally 40 epochs in ACRP stage, as the classification results of the real images converge quickly, we stop CRP Sampling (re-classification) after 10 epochs, so the GAN can focus on improving the quality of image generation.

For the dataset of Tiny ImageNet, we picked 10 classes for the MIC-GANs training, which are `goldfish', `black widow', `brain coral', `golden retriever', `monarch', `beach wagon', `beacon', `bullet train', `triumphal arch', `lemon'.

\begin{table}[tb]
  \begin{center}
  \begin{tabular}{|c|c|c|c|}
  \hline
  & training Q & sampling & training GAN \\
  \hline
  DCGAN & 0.5mins & 2.6mins & 4.5mins \\
  \hline
  StyleGAN2 & 0.5mins & 2.6mins & 15mins \\
  \hline
  \end{tabular}
  \caption{Training time distribution for one epoch on MNIST.}
  \vspace{-0.7cm}
  \label{tab:time}
  \end{center}
\end{table}

\begin{table}[tb]
  \begin{center}
  \begin{tabular}{|c|c|c|c|c|c|}
  \hline
  epochs & 1 & 5 & 9 & 13 & 19  \\
  \hline
  purity & 0.839 & 0.908 & 0.911 & 0.927 & 0.929 \\
  \hline
  \end{tabular}
  \caption{Purity vs sampling epochs on MNIST with $K=15$.}
  \vspace{-0.7cm}
  \label{tab:purity}
  \end{center}
\end{table}

\section{Quantitative Results}

Table.~\ref{tab:time} shows the training time distribution for one epoch on MNIST dataset. We find that the sampling in learning the prior is not the most time-consuming component, while the training of the GANs itself dominates the training time. And the situation is similar on all datasets. 

In Table.~\ref{tab:purity}, we show the relationship between the purity and sampling epochs on the MNIST dataset. We find that the purity converges quickly in the first few epochs (similar on other datasets). So we stop the CRP sampling after 10 epochs and use the stable classification results for GAN training. 

\section{Generation Results}
\begin{figure*}[htp]
  \centering
  \includegraphics[width=1.0\linewidth]{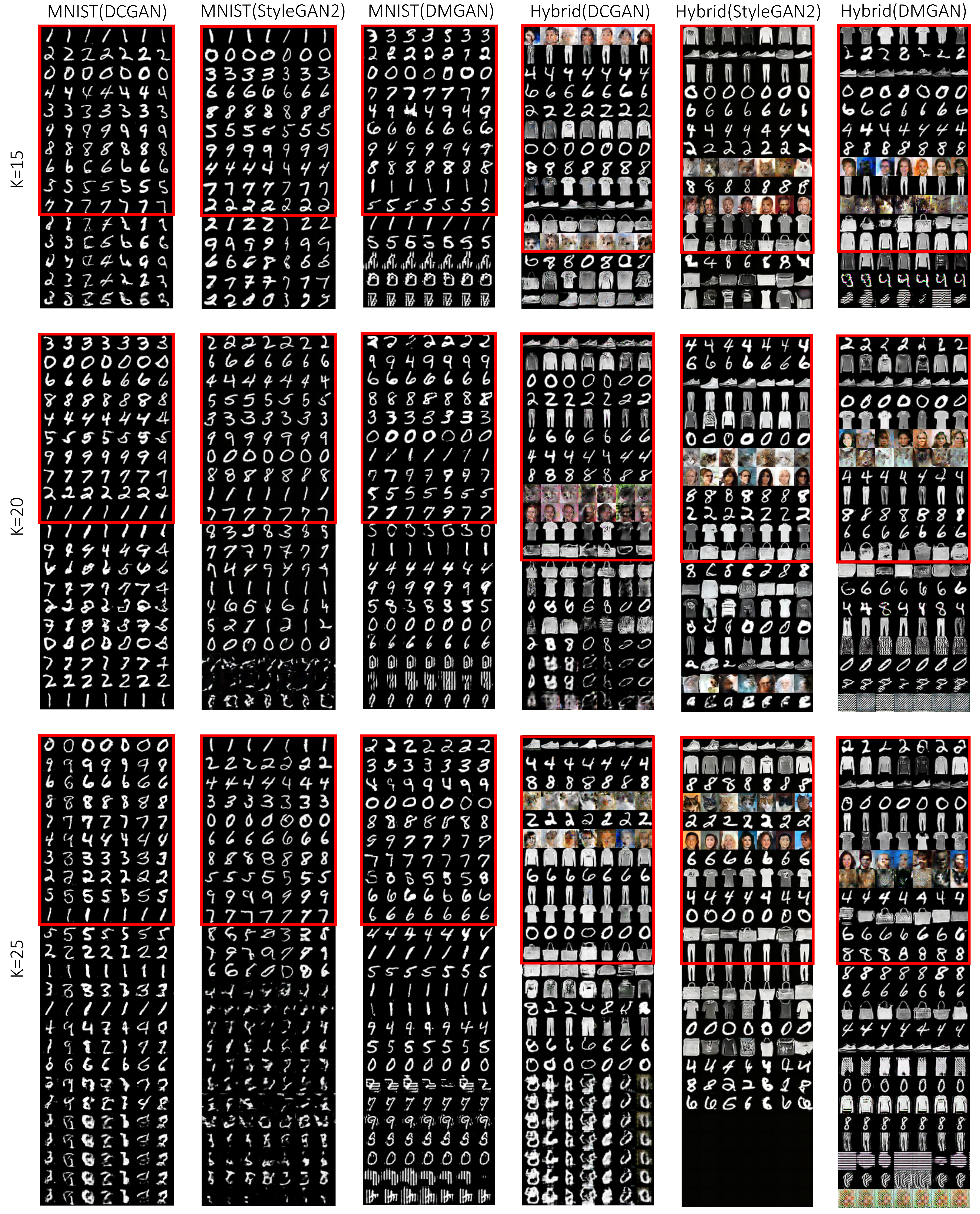}
  \caption{Our results on the MNIST and Hybrid dataset using DCGAN and StyleGAN2 with different $K$s, compared to DMGAN. Each row is generated from a mode, and the rows are sorted by $\alpha$s. The red boxes mark the top $n$ modes in the results, where $n=10$ for MNIST and $n=12$ for Hybrid.}
  \vspace{-0.0cm}
\label{fig:supp_ours}
\end{figure*}

Figure~\ref{fig:supp_ours} visualizes the MNIST and Hybrid results of our method and DMGAN~\cite{khayatkhoei2018disconnected}. We can find that even using different $K$s, our method can provide stable results, which demonstrates the ability of unsupervised clustering of our method. DMGAN achieves a similar effect as ours, but it often learns mixed modes, e.g., the confusion between '4' and '9' on MNIST. Furthermore, our method is flexible with the architecture of GAN without compromising the training speed much, which means that we can employ complex GANs such as StyleGAN2 on complex datasets. However, it will be prohibitively expensive for DMGAN to achieve the same because DMGAN requires $K$ generators for $K$ modes, while MIC-GANs only require $K$ latent codes.

\begin{figure*}[htp]
  \centering
  \includegraphics[width=0.95\linewidth]{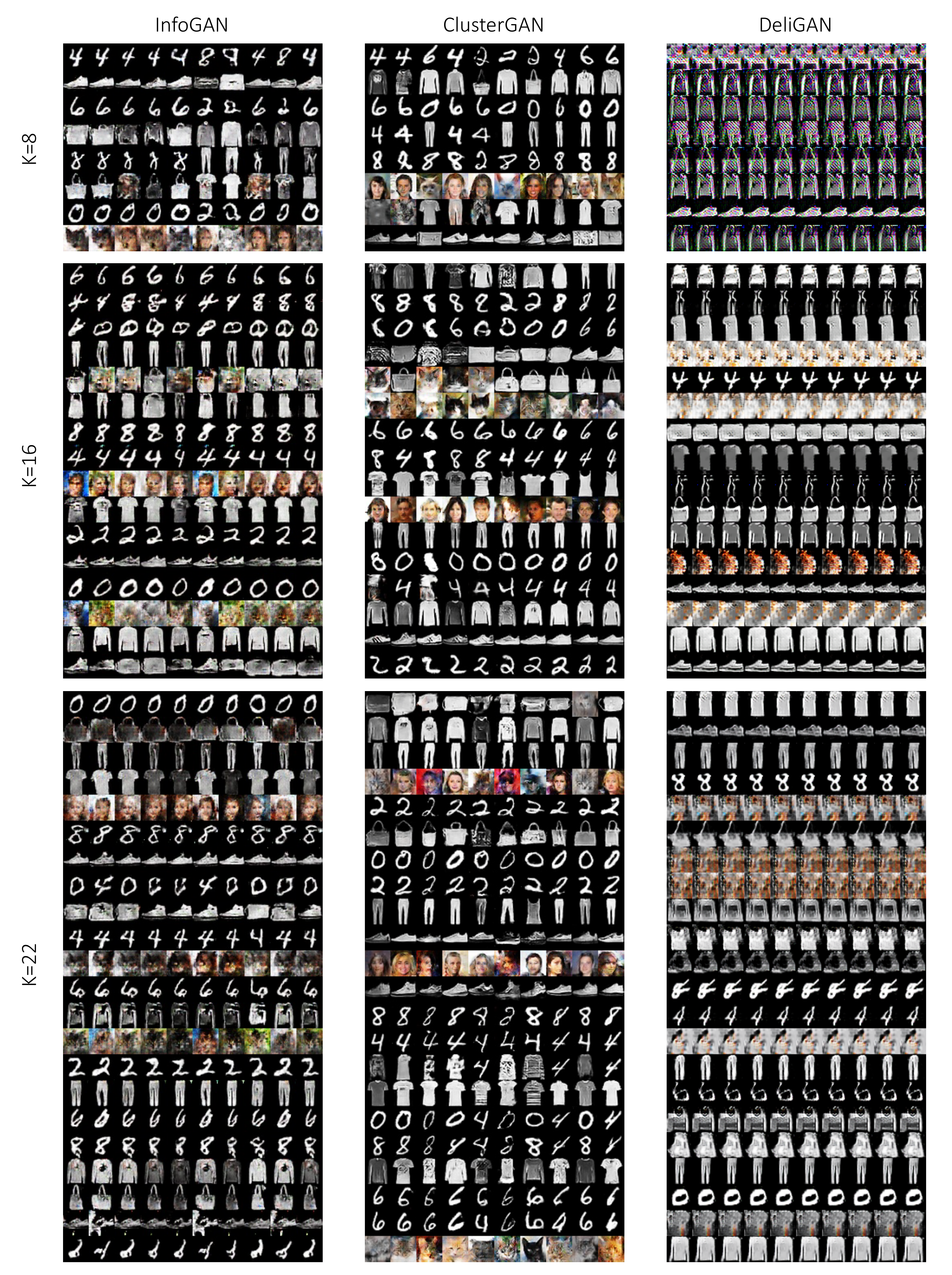}
  \caption{Results of InfoGAN, ClusterGAN and DeliGAN on the Hybrid dataset with different $K$s. Each row is generated from a mode.}
  \vspace{-0.3cm}
\label{fig:supp_other}
\end{figure*}

\begin{figure*}[htp]
  \centering
  \includegraphics[width=0.95\linewidth]{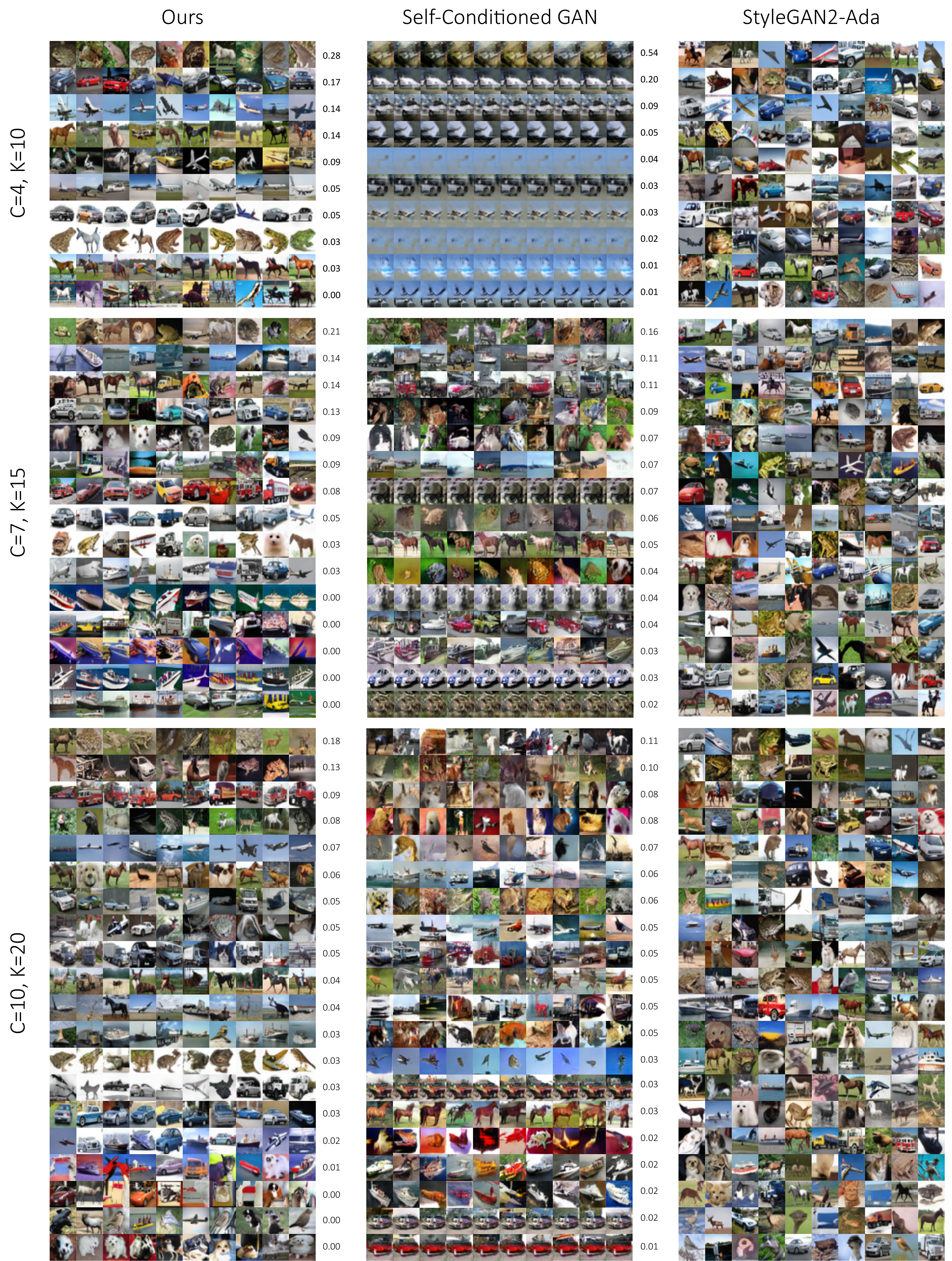}
  \caption{Results of Our method, Self-Conditioned GAN and StyleGAN2-Ada on the CIFAR with different $C$s and $K$s. ‘$C$’ means the ground-truth class number in the dataset. Note that StyleGAN2-Ada is trained without conditions. For our method and Self-Conditioned GAN, each row is generated from a mode, and the number on the right of each row indicates the distribution of the mode.}
  \vspace{-0.3cm}
\label{fig:cifar}
\end{figure*}

\begin{figure*}[htp]
  \centering
  \includegraphics[width=0.75\linewidth]{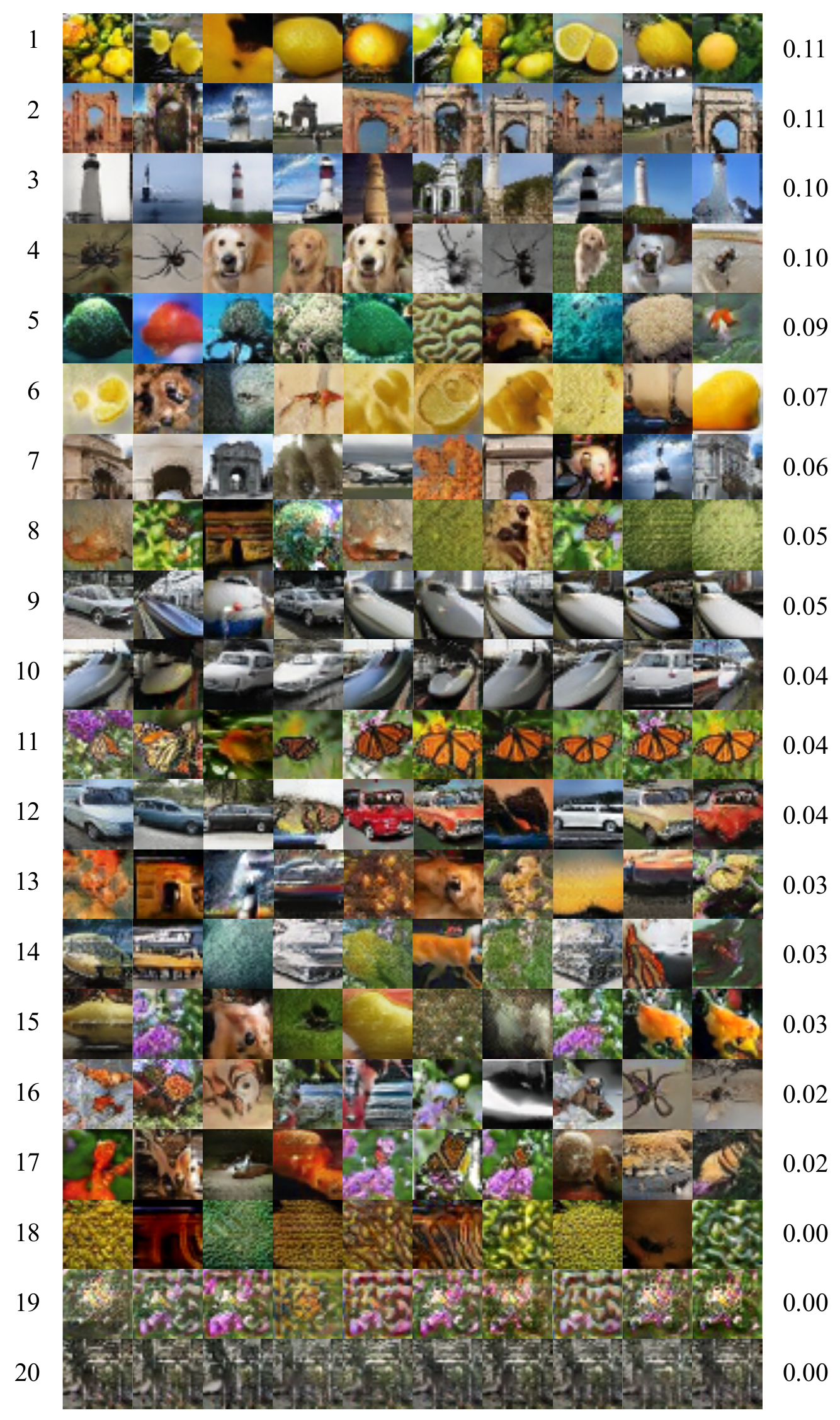}
  \caption{Results of Our method on the Tiny Imagenet with different $C=10$s and $K=20$s. Each row is generated from a mode,. The number on the left of each row indicates the line number and on the right indicates the weight of the mode.}
  \vspace{-0.3cm}
\label{fig:imagenet}
\end{figure*}

Figure~\ref{fig:supp_other} shows the results of InfoGAN~\cite{infoGAN16}, ClusterGAN~\cite{mukherjee2019clustergan} and DeliGAN~\cite{deliGAN17} on Hybrid with different $K$s. Obviously, these methods fail to perform correct clustering when the ground-truth $K=12$  is unknown and the best way is to make multiple guesses. However, when $K<12$, there will be mixed modes; when $K>12$, there will be repetitive modes, as well as mixed modes. This shows that these methods either cannot produce good results or require a large number of guesses in the absence of the ground-truth, while MIC-GANs can generate satisfying results in one run.

Figure~\ref{fig:cifar} shows the generation results of our method, Self-Conditioned GAN~\cite{liu_self_2020} and StyleGAN2-Ada~\cite{Karras2020ada} on CIFAR with different $C$s and $K$s. CIFAR is a difficult dataset for generation, and it is an even more challenging dataset for conditioned generation based on unsupervised clustering. In our method, some modes can generate images that are from clear-cutting single classes, e.g., `automobile', `airplane', `horse'. In other cases, images generated from one mode consist of images from two or more classes. This reflects the fact that images can be clustered based on different criteria. This sometimes leads to different classification results between MIC-GANs and human labels. For example, images can be classified according to the colors or shapes or semantics. While human labels in CIFAR are primarily based on semantics (object identities), it is normal that MIC-GANs at times generate images from one mode that match several ground-truth classes. Nevertheless, we can still find some interesting similarities among the images generated from one mode. In addition, MIC-GANs improve the generation quality in general with lower FID scores shown in the paper. We also find that Self-Conditioned GAN suffers from mode collapse in several modes and the problem gets worse when $K$ is small. StyleGAN2-Ada is able to generate images with diversities but ours still achieve better FID scores.

Figure~\ref{fig:imagenet} shows the generation results of our method on the Tiny Imagenet dataset. 
Without any class supervision, our algorithm still generates several reasonable conditional results. For example, line 1, 2, 3, 5, 11, 12 correctly produce the images of lemon, triumphal arch, beacon, brain coral, monarch and beach wagon, while several modes generate the mixtures of classes, like line 4 and 7. Another interesting observation is that line 9 mostly generates bullet trains facing the right while line 10 genearates buleet trains facing the left. Tiny Imagenet is a difficult dataset, so the generation is less ideal on some modes, but still covers most of them.

\end{appendix}

\end{document}